\documentclass[10pt,twocolumn,letterpaper]{article}

\usepackage{cvpr}              

\usepackage[accsupp]{axessibility}
\usepackage[normalem]{ulem}
\usepackage{graphicx}
\usepackage{amsmath}
\usepackage{amssymb}
\usepackage{booktabs}
\usepackage{tabularx}
\usepackage{tablefootnote}
\usepackage[pagebackref,breaklinks,colorlinks]{hyperref}
\usepackage[capitalize]{cleveref}
\usepackage{multirow}
\usepackage{booktabs}
\crefname{section}{Sec.}{Secs.}
\Crefname{section}{Section}{Sections}
\Crefname{table}{Table}{Tables}
\crefname{table}{Tab.}{Tabs.}

\begin{document}

\title{GCFAgg: Global and Cross-view Feature Aggregation for  Multi-view Clustering}

\author{Weiqing Yan$^{1,5}$, Yuanyang Zhang$^1$, Chenlei Lv$^2$, Chang Tang$^{3}$\thanks{Corresponding author(tangchang@cug.edu.cn)}, Guanghui Yue$^4$, Liang Liao$^5$, Weisi Lin$^5$\\
\small{$^1$School of Computer and Control Engineering, Yantai University, Yantai 264005, China }\\
\small{$^2$College of Computer Science and Software Engineering, 
Shenzhen University, Shenzhen 518060, China}\\
\small{$^3$ School of Computer, China University of Geosciences, Wuhan 430074, China}\\ 
\small{$^4$ School of Biomedical Engineering, Health Science Center, Shenzhen University, Shenzhen 518060, China}\\
\small{$^5$ School of Computer Science and Engineering,  Nanyang Technological University, 639798, Singapore.}}


\maketitle
\begin{abstract}

Multi-view clustering can partition data samples into their categories by learning a consensus representation in unsupervised way and has received more and more attention in recent years. However, most existing deep clustering methods learn consensus representation or view-specific representations from multiple views via view-wise aggregation way, where they ignore structure relationship of all samples.   
In this paper, we propose a novel multi-view clustering network to address these problems, called Global and Cross-view Feature Aggregation for Multi-View Clustering (GCFAggMVC). Specifically, the consensus data presentation from multiple views is obtained via cross-sample and cross-view feature aggregation, which fully explores the complementary of similar samples. Moreover, we align the consensus representation and the view-specific representation by the structure-guided contrastive learning module, which makes the view-specific representations from different samples with high structure relationship similar.  The proposed module is a flexible multi-view data representation module, which can be also embedded to the incomplete multi-view data clustering task via plugging our module into other frameworks. 
Extensive experiments
show that the proposed method achieves excellent performance in both complete multi-view data clustering
tasks and incomplete multi-view data clustering tasks.

\end{abstract}
\section{Introduction}
With the rapid development of  informatization, data is often collected by various social media or views. For instance, a 3D object can be described from different angles; a news event is reported from different sources; and an image can be characterized by different types of feature sets, e.g., SIFT, LBP, and HoG. Such an instance object, which is described from multiple views, is referred to as multi-view data. Multi-view clustering (MVC)~\cite{ chao2021survey}, i.e., unsupervisedly fusing the multi-view data to aid differentiate crucial grouping, is a fundamental task in the fields of data mining, pattern recognition, etc, but it remains a challenging problem. 
\par
Traditional multi-view clustering methods \cite{Chem2022review} include matrix decomposition methods, graph-based multi-view  methods, and subspace-based multi-view  methods. The goal of these methods is to obtain a high-quality consensus graph or subspace self-representation matrix by various regularization constraints in order to improve the performance of clustering.  However, most of them directly operate on the original multiview  features or specified kernel features, which usually include noises and redundancy information during the collection or kernel space selection processes, moreover harmful to the clustering tasks. 
\par

Deep neural networks  have demonstrated excellent performance in data feature representation for many vision tasks. Deep clustering methods also draw more attention to researchers~\cite{ du2021deep,abavisani2018deep,zhou2020end,xu2021deep,trosten2021reconsidering, xu2022multi}. These methods efficiently learn the feature presentation of each view using a view-specific encoder network,  and  fuse these learnt representations from all views to obtain a consensus representation that can be divided into different categories by a clustering module. To reduce the influence of view-private information on clustering, these methods designed different alignment models. For example, some methods align the representation distributions or label distributions from different views by KL divergence \cite{hershey2007approximating}. They might be hard to distinguish between clusters, since a category from one view might be aligned with a different category in another view. Some methods align the representation from different views by contrastive learning. Despite these models have achieved significant improvement in MVC task, the following issues still exist: 1) Almost all existing deep MVC methods (such as \cite{trosten2021reconsidering,tang2022deep,xu2022deep}) are based on view-wise fusion models, such as weighted-sum fusion of all views or concatenating fusion of all views, which makes it difficult to obtain discriminative consensus representations from multiple views,  since a view or several views of a sample might contain too much noise or be missing in the collection process. 
2) These alignment methods (such as \cite{xu2022multi,tang2022deep}) based on contrastive learning  usually distinguish the positive pair and negative pair from the sample-level.  That is, they make inter-view presentations from the same sample as positive pair, and makes view representations from different samples as negative pair (including view representations from different samples in the same cluster),  whereas it might be conflict with the clustering objective, where these representations of samples from the same cluster should be similar to each other.

In the paper,  a novel multi-view representation learning framework for clustering is proposed to alleviate the above problems. Motivated by the insight that the representations of samples from the same category are typically similar, we can learn consensus data representation from multiview data by other samples with a high structure relationship, moreover, in contrastive learning, we should increase the similarity of view representations from the same cluster, not only from the same sample, which is beneficial clustering tasks. 
To accomplish this, we first learn the view-specific representations to reconstruct the original data by leveraging the autoencoder model.  Then, we design a global and cross-view feature aggregation module, which is capable of learning a global similarity relationship among samples, and obtaining a consensus representation based on the global similarity relationship of all samples.  Furthermore, we leverage the learnt global structure relationship and consensus representation to establish the consistency with view-specific representations by contrastive learning, which minimizes the similarity between the representations with low structure relationship. Compared with previous work, our contributions are listed as follows:

\begin{itemize}
\par
\item We propose a novel Global and Cross-view Feature Aggregation network framework for Multi-View Clustering (GCFAggMVC), which is able to fully explore the complementary of similar samples and addresses the problem of negative pairs from the different samples in the same cluster having low similarity score.
\item Different from previous methods, we design a global and cross-view feature aggregation module, which integrates the transformer structure to learn the global structure relationship from different feature spaces, and then obtains the consensus representation based on the learnt global relationship, which fully exploits the complementary information of similar samples, thereby reduce the impact of noise and redundancy or sample missing among different views.  Moreover, we align the consensus representation and the view-specific representation by our global structure-guided contrastive learning module, which makes the representations of similar samples with highly structure relationship similarity.
\item The proposed module is flexible multi-view data representation module, which can be also applied to incomplete multi-view data clustering tasks by plugging our module into the framework of other methods as the consensus representation of a sample with missing view data can be enhanced by these samples with high structure relationships. Experiments show that the proposed method achieves not only excellent performance in complete multi-view clustering tasks, but also works well in incomplete multi-view clustering tasks.

\end{itemize}

\par
\begin{figure*}[!ht]
\centering
\includegraphics[width=0.9\linewidth]{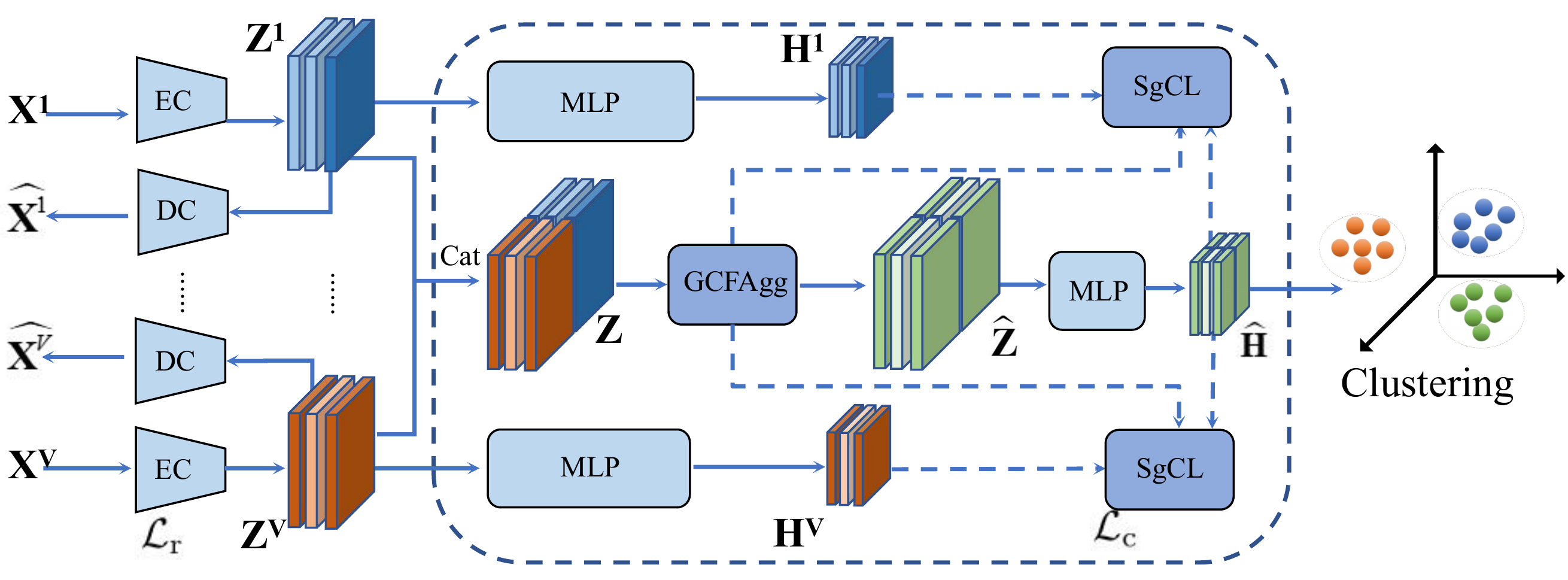}
\vspace{-2mm}
\caption{The overall framework. Our module includes global and cross-view feature aggregation module (GCFAgg) and structure-guided multiview contrastive learning module (SgCL). The former learns a consensus representation via considering global structure relationship among samples, which fully explores the complementary of similar samples. The latter integrates the learnt global structure relationship and consensus representation to contrastive learning, which makes data representations in the same cluster similar and addresses the aforementioned second issue in the introduction. Note that, EC: Encoder; DC: Decoder; Cat: Concatenation; MLP: Multi-Layer Perception.
  }
\label{fig:framework}
\vspace{-2.5mm}
\end{figure*}
\section{RETATED WORK}
In the existing representation learning-based MVC family \cite{Chem2022review}, there are mainly two kinds of methods, i.e., shallow representation learning-based method and deep representation learning-based  method. The following section, a brief review will be introduced.

\subsection{Shallow representation learning-based method}

Shallow representation learning-based multi-view clustering methods \cite{Chem2022review} are divided into the  two categories, multi-view graph clustering and multi-view subspace clustering. 

For multiview graph clustering~\cite{hu2020multi,jing2021learning,fang2022structure,pan2021multi,nie2017self,lin2021multi},  they generally do the following steps: construct view-specific graphs, and then obtain a fusion graph by different regularization terms from multiple view-specific graphs, last, produce the clustering results by spectral clustering or graph-cut methods or other algorithms. 

For subspace-based multiview clustering \cite{gao2015multi,cao2015diversity,wang2017exclusivity, kang2020large, lv2021multi,sun2021scalable}, they learn a consensus subspace self-representation matrix from each view, and partition the data points into different subspaces by applying different regularization terms on the consensus self-representation matrix.  To reducing the algorithm complexity, some methods \cite{kang2020large, sun2021scalable} computed similarity graph or self-expression depending on the bipartite graph between samples and anchors instead of  all samples-based graph methods. These methods directly learn graph or self representation in the original feature space in which redundant and noisy features of samples are inevitably mixed. Some methods \cite{el2022consensus,li2021consensus,liu2021multiview} map raw feature space to high-dimensional feature space by kernelization method, however, these methods assume that the raw data can be represented by one or fixed few kernel matrices, which might be insufficient to describe complex data. 

 

\subsection{Deep representation learning-based method}

Motivated by the promising progress of deep learning in  powerful feature
transformation ability, many recent works have been focused on the deep representation learning-based multi-view clustering. Specifically, these methods use the deep neural network  to model  the non-linear parametric mapping functions, from which the embedded feature representation canbe learnt the manifold structure by the non-linearity properties.  

Some deep multi-view clustering methods \cite{ li2019deep,zhou2020end,li2021multi,wang2022adversarial} use adversarial training to  learn the latent feature representations and align distributions of hidden representations from different views. Zhou et al. \cite{zhou2020end} leveraged the attention mechanism  to obtain a weight value for each view, and obtained a consensus representation by weight summation of all view-wise presentations. Wang et al. \cite{wang2022adversarial} obtained the consensus representation by a weight summation and $l_{1,2}$-norm constraint.
Contrastive learning (CL) is able to align representations from different views at the sample level, further facilitating the alignment of label distributions as well.
These CL methods~\cite{chen2020simple,tian2020contrastive, lin2021completer, pan2021multi, ke2021conan, trosten2021reconsidering, xu2022multi}  have exceeded the previous  distribution alignment methods to multi-view clustering, yielding SOTA clustering performance on several multi-view datasets.

\section{PROPOSED METHOD}
The proposed framework is shown in Fig. \ref{fig:framework}. In this paper, a multi-view data, which includes n samples with V views, is denoted as $\{\mathbf{X}^{v}=\{\mathbf{x}_{1}^{v};...;\mathbf{x}_{N}^{v}\}\in \mathbb{R}^{N \times D_{v}}\}_{v=1}^V$, where $D_v$ is the feature dimension in the $v$-th view. 
\subsection{Multi-view Data Reconstruction}
Original multi-view data usually contains redundancy and random noise,  we need to first learn representative feature representations from the original data features. In particular, autoencoder\cite{song2018self,hinton2006reducing} is a widely used unsupervised model that projects  original data features into a designed feature space. Specifically, for the \emph{v}-th view, let $f_{\theta^{v}}^{v}(.)$ represent the encoder nonlinear function. In the encoder, the network learns the low-dimensional features as follows:
\begin{equation}
\label{eq:encoder}
\begin{aligned}
\mathbf{z}_{i}^{v}=f_{\theta^{v}}^{v}\left(\mathbf{x}_{i}^{v}\right), 
\end{aligned}
\end{equation} where $\mathbf{z}_{i}^{v} \in \mathbb{R}^{d_v}$ is the embedded data representation  in ${d_v}$-dimensional feature space of the $\textit{i}$-th sample from the \emph{v}-th view $\mathbf{x}_{i}^{v}$.

The decoder reconstructs the sample by the data representation $\mathbf{z}_{i}^{v}$. Let $g_{\eta^{v}}^{v}(.)$ represent the decoder function, in decoder part, the reconstructed sample $\mathbf{\hat{x}}_{i}^{v}$ is obtained by decoding $\mathbf{z}_{i}^{v}$:
\begin{equation}
\label{eq:decoder}
\begin{aligned}
\hat{\mathbf{x}}_{i}^{v}=g_{\eta^{v}}^{v}\left(\mathbf{z}_{i}^{v}\right)=g_{\eta^{v}}^{v}\left(f_{\theta^{v}}^{v}\left(\mathbf{x}_{i}^{v}\right)\right)
\end{aligned}
\end{equation}
Let $\mathcal{L}_{r}$ be the reconstruction loss from input to output $\hat{\mathbf{X}}^{v}=\{\hat{\mathbf{x}}_{1}^{v};...;\hat{\mathbf{x}}_{n}^{v}\} \in \mathbb{R}^{n \times D_{v}}$, $n$ denotes the number of
samples in a batch. The reconstruction loss is formulated as:
\begin{equation}
\label{eq:reconstruction loss}
\begin{aligned}
\mathcal{L}_{\mathrm{r}}=\sum_{v=1}^{V}\mathcal{L}_{\mathrm{r}}^{v}=\sum_{v=1}^{V}\left\|\mathbf{X}^{v}-\hat{\mathbf{X}}^{v}\right\|_{2}^{2}\\
=\sum_{v=1}^{V}\sum_{i=1}^{n}\left\|\mathbf{x}_i^{v}-g_{\eta^{v}}^{v}\left(\mathbf{z}_{i}^{v}\right)\right\|_{2}^{2}
\end{aligned}
\end{equation}

\subsection{Global and Cross-view Feature Aggregation}
In Eq. (3), to reduce the loss of reconstruction, the learned data representation $\mathbf{z}_{i}^{v}$  usually contains much view-private information, which may be meaningless or even misleading. They might result in poor clustering quality if they are fused by previous view-wise fusion methods.  

\begin{figure}[h]
\centering
\includegraphics[width=8cm]{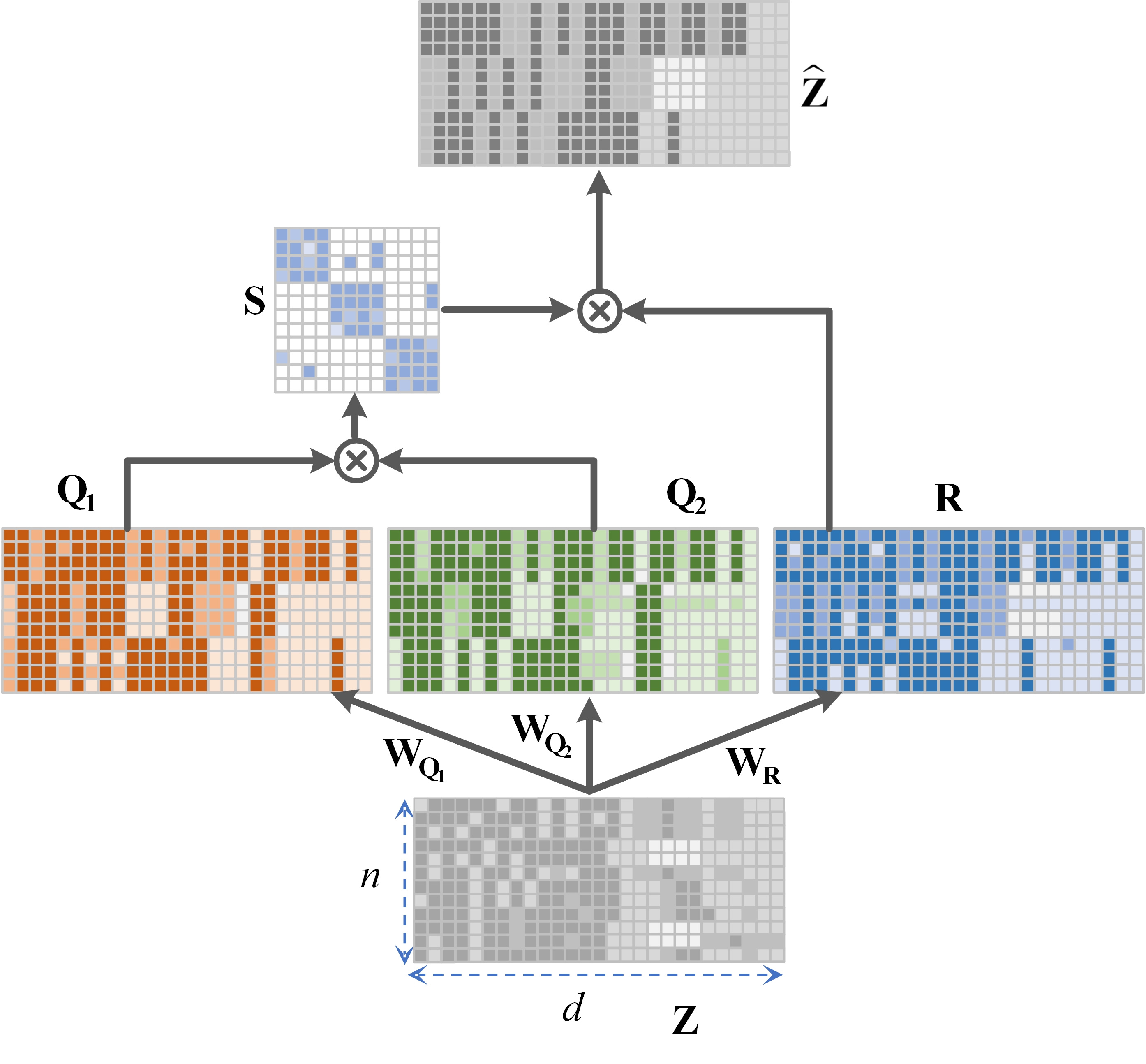}
\vspace{-2mm}
\caption{Global and Cross-view Feature Aggregation module . ${\mathbf{W}_{Q_1},\mathbf{W}_{Q_2},\mathbf{W}_{R}}$ is used to achieve feature space transformation of cross-view, and the $\mathbf{S}$ is used to obtain global structure relationship among samples.}
\vspace{-2mm}
\label{fig:transformerencoder}
\end{figure}

Since the representations of different samples in the same cluster are similar, the consensus representation of a sample from different views should  be enhanced by those samples with  strong structure relationships, not just the weight-sum of representation of different views from the same sample. 
Therefore, to obtain consensus  multi-view data representation, we learn the global structure relationship among samples and use it to obtain the consensus data representation. The module is shown in Figure 2.  
Specifically, we firstly concatenate all view-specific representations $\mathbf{Z}^{v}$ extracted by the encoder together to get $\mathbf{Z}$ is formulated as:
\begin{equation}
\label{eq:concatenation operation}
\begin{aligned}
\mathbf{Z} =\left[\mathbf{Z}^{1}, \mathbf{Z}^{2}, \ldots, \mathbf{Z}^{V}\right]
\end{aligned}
\end{equation}
where $\mathbf{Z}^{v}=\left[\mathbf{z}_1^{v}; \mathbf{z}_2^{v}; \ldots; \mathbf{z}_n^{v}\right] \in \mathbb{R}^{n \times d_v} $, $\mathbf{z}_i^{v}\in \mathbb{R}^{1 \times d_v}$, $\mathbf{Z} \in \mathbb{R}^{n \times d}$, $d=d_v*V$. Here, we set each sample as a token, set linear encoding and position encoding.
\par

Inspired by the idea of the transformer attention mechanism\cite{vaswani2017attention}, we map $\mathbf{Z}$ into different feature spaces by the $\mathbf{W}_{R}$   to achieve the cross-view fusion of all views, i.e.,
\begin{equation}
\label{eq:cross}
\left[ {\begin{array}{*{20}{c}}
{{{\bf{R}}_{1:}}}\\
{{{\bf{R}}_{2:}}}\\
 \vdots \\
{{{\bf{R}}_{n:}}}
\end{array}} \right] = \left[ {\begin{array}{*{20}{c}}
{{\bf{z}}_1^1}&{{\bf{z}}_1^2}& \cdots &{{\bf{z}}_1^V}\\
{{\bf{z}}_2^1}&{{\bf{z}}_2^2}& \cdots &{{\bf{z}}_2^V}\\
 \vdots & \vdots & \ddots & \vdots \\
{{\bf{z}}_n^1}&{{\bf{z}}_n^2}& \cdots &{{\bf{z}}_n^V}
\end{array}} \right]\left[ {\begin{array}{*{20}{c}}
{{{\bf{W}}_{R1:}}}\\
{{{\bf{W}}_{R2:}}}\\
 \vdots \\
{{{\bf{W}}_{RV:}}}
\end{array}} \right]
\end{equation}
that is ${\bf{R}}_{j:}=\sum\limits_{k = 1}^V {\bf{z}}_j^k{\bf{W}}_{Rk:}$. Similarity, the $\mathbf{Q}_1$ and $\mathbf{Q}_2$ is obtained by  $\mathbf{W}_{Q_1},\mathbf{W}_{Q_2}$,  i.e., 
\begin{equation}
\label{eq:concatenation q1}
\begin{aligned}
\mathbf{Q}_1 =\mathbf{Z}\mathbf{W}_{Q_1};
\mathbf{Q}_2 =\mathbf{Z}\mathbf{W}_{Q_2};
\end{aligned}
\end{equation}
where $\mathbf{Q}_1\in \mathbb{R}^{n \times d}$, $\mathbf{Q}_2\in \mathbb{R}^{n \times d}$. Here,  we use the  matrix $\mathbf{W_Z}=\{\mathbf{W}_{Q_1},\mathbf{W}_{Q_2},\mathbf{W}_{R}\}$ to denote the parameters in this module. 

The structure relationship among samples is denoted as:
\vspace{-0.5em}
\begin{equation}
\label{eq:s}
\begin{aligned}
\mathbf{S}=\operatorname{softmax}\left(\frac{\mathbf{Q}_1\mathbf{Q}_2^{T}}{\sqrt{d}}\right). 
\end{aligned}
\end{equation}
The learned representation matrix $\mathbf{R}$ is enhanced by the global structure relationship matrix  $\mathbf{S}$. That is, the data representation of the sample can be enhanced by other samples with high correlation, which makes  these data representations of samples in the same cluster be similar. 
\begin{equation}
\label{eq:fully con}
\begin{aligned}
\widehat {{{\bf{z}}_i}} = \sum\limits_{j = 1}^n {{{\bf{S}}_{ij}}{\bf{R}}_{j:}}; { }{ }
\hat{\bf{Z}} =\left[\widehat {{{\bf{z}}_1}}; \widehat {{{\bf{z}}_2}}; \ldots; \widehat {{{\bf{z}}_n}}\right]
\end{aligned}
\end{equation}
where ${\bf{R}}_{j:}\in \mathbb{R}^{1 \times d}$ is the $j$-th row elements of $\mathbf{R}$, denotes the $j$-th sample representation, ${\bf{S}}_{ij}$ denotes the relationship between the $i$-th sample  and the $j$-th sample,  $\hat{\bf{Z}}\in \mathbb{R}^{n \times d}$. Since $\hat{\bf{Z}}$ is learnt from the concatenation of all views $\bf{Z}$, it usually contains redundancy information.  Next, the output is passed through the fully connected nonlinear and linear layer to eliminate the redundancy information. The expression is described as the following equation: 
\begin{equation}
\label{eq:fully con}
\begin{array}{l}
\widehat {\bf{H}} = {\bf{W}_3}\left( {\max (0,(\bf{Z}+\hat{\bf{Z}}){W_1} + {b_1}){W_2} + {b_2}} \right) + {\bf{b}_3}
\end{array}
\end{equation}
where we use ${{\bf{W}}_{\hat{\bf{H}}}}$ to denote these parameters in the layers.

\subsection{Structure-guided Contrastive Learning}

The learnt consensus representation $\hat{\bf{H}} $ is enhanced by global structure relationship of all samples in a batch, these data consensus representations from different views of samples in the same cluster are similar. Hence,  the consensus representation $\hat{\bf{H}} $  and view-specific representation $\mathbf{H}^{v}$ from the same cluster should be mapped close together. 
Inspired by contrastive learning methods \cite{chen2020simple},  we set the consensus representation and view-specific representation from different views from the same sample as positive pairs. However, if other pairs are directly set as negative pairs, it might lead to the inconsistency of these represents from different samples in the same cluster, which is conflict with clustering objective.  Hence, we design a structure-guided multiview contrastive learning module. Specifically, 
we introduce cosine distance to compute the similarity between consensus presentation $\mathbf{\hat{H}}$ and view-specific presentation $\mathbf{H}^{v}$:
\begin{equation}
\label{eq:sim}
\begin{aligned}
C\left(\mathbf{\hat{H}}_{i:}, \mathbf{H}_i^{v}\right)=\frac{\mathbf{\hat{H}}_{i:}^{T}\mathbf{H}_{i:}^{v}}{\left\|\mathbf{\hat{H}}_{i:}\right\|\left\|\mathbf{H}^{v}_{i:}\right\|}
\end{aligned}
\end{equation}
The loss function of structure-guided multi-view contrastive learning can be defined as:
\begin{equation}
\label{eq:lc}
\begin{aligned}
\mathcal{L}_{\mathrm{c}}=-\frac{1}{2 N} \sum_{i=1}^{N} \sum_{v=1}^{V} \log \frac{e^{\operatorname{C}\left(\mathbf{\hat{H}}_{i:}, \mathbf{H}_{i:}^{v}\right) / \tau}}{\sum_{j=1}^{N} e^{(1-\mathbf{S}_{ij})\operatorname{C}\left(\mathbf{\hat{H}}_{i:}, \mathbf{H}_{j:}^{v}\right) / \tau}-e^{1 / \tau}}
\end{aligned}
\end{equation}
where $\tau$ denotes the temperature parameter, $\mathbf{S}_{ij}$ is from Eq. (\ref{eq:s}). In this equation, when the smaller this  $\mathbf{S}_{ij}$ value is, the bigger the $\operatorname{C}\left(\mathbf{\hat{H}}_{i:},\mathbf{H}_{j:}^{v}\right)$. 
In other words, when the structure relationship $\mathbf{S}_{ij}$ between the $i$-th and $j$-th sample is low (not from the same cluster), their corresponding representations are inconsistent; otherwise, their corresponding representations are consistent, which solves the aforementioned second issue in the introduction. 

In the proposed framework, the loss in our network consists of two parts:
\begin{equation}
\label{zong}
\begin{aligned}
\begin{array}{l}
{\cal L} = {{\cal L}_{\rm{r}}} + \lambda {{\cal L}_{\rm{c}}}\\
\begin{array}{*{20}{c}}
{}&{ = {{\cal L}_{\rm{r}}}\left( {\{ {{\bf{X}}^v},{{\widehat {\bf{X}}}^v}\}_{v=1}^V ;\{ {\eta ^v},{\theta ^v}\}_{v=1}^V } \right) + }
\end{array}\\
\begin{array}{*{20}{c}}
{}&{}
\end{array}{\lambda{\cal L}_{\rm{c}}}\left( {\{ {\bf{Z}},\widehat {\bf{H}},{{\bf{H}}^v}\} _{v=1}^V;\{ {{\bf{W}}_{\bf{Z}}},{{\bf{W}}_{\widehat {\bf{H}}}},{\bf{W}}_{{\bf{H}}^v},{\theta ^v}\}_{v=1}^V } \right)
\end{array}
\end{aligned}
\end{equation}
where ${\cal L}_{\rm{r}}$ is the reconstruction loss from input ${\bf{X}}^v$ to the output $\widehat {\bf{X}}^v$ by the embedding representation matrix ${\bf{Z}}^v$, ${\eta ^v},{\theta ^v}$ denote the parameters of encoder and decoder, this loss is used to avoid the model collapse. ${\cal L}_{\rm{c}}$ is the consistent data presentation loss by contrastive learning between structure-enhanced consensus data representation  $\widehat {\bf{H}}$ and view-specific data representation ${\bf{H}}^v$. 

Different from previous contrastive learning methods, which align inter-view representations, our method has the following advantages: 
 1) we add the structure relationship $\mathbf{S}$ (Eq. (\ref{eq:s})) obtained by feature fusion to negative pairs, which ensures that we only minimize the similarity between the view-specific representation and the consensus representation from different samples with low structure relationship. 2) for the positive
pairs, the proposed method aligns the learnt consensus representation and view-specific representation, which makes
the representations of positive pair with high structure relationship be more similar, since the consensus presentation is enhanced with by other samples with high correlation. The
proposed method breaks the limitations of previous CL
at sample level.

\subsection{Clustering module}

For the clustering module, we take the k-means \cite{mackay2003information,bauckhage2015k} to obtain the clustering results for all samples. Specifically, the learnt consensus representation $\widehat {\bf{H}}$ is factorized as follows:
\begin{equation}
\label{clustering1}
\begin{array}{l}
\begin{array}{*{20}{c}}
{\mathop {\min }\limits_{{\bf{U,V}}} }&{{{\left\| {\widehat {\bf{H}} - {\bf{UV}}} \right\|}^2}}
\end{array}\\
s.t.{\bf{U1}} = {\bf{1}},{\bf{U}} \ge {\bf{0}}
\end{array}
\end{equation}
where ${\bf{U}} \in \mathbb{R}^{n \times k} $ is cluster indicator matrix; ${\bf{V}} \in \mathbb{R}^{k\times d}$ is the center matrix of clustering.      

\subsection{Optimization}
The model consists of multiple encoder-decoder modules and multiple MLP layers. The model is optimized by a mini-batch gradient descent algorithm. Firstly, the autoencoders are initialized by Eq. (\ref{eq:reconstruction loss}), and structure-guided contrastive learning are conducted to obtain the consensus  representation by Eq. (\ref{zong}). At last, the cluster labels are obtained by Eq. (\ref{clustering1}).

\section{Experiments}
\subsection{Experimental Settings}
We evaluate our models on 13 public multi-view datasets with different scales (see Table \ref{tab:Datasets}). For the evaluation metrics, three metrics, including accuracy (ACC), normalized mutual information (NMI), and Purity (PUR). 
\begin{table}[!ht]
\centering
\renewcommand\tabcolsep{8pt}
\caption{Description of the multiview datasets.}
\begin{tabular}{cccc} 
\toprule
Datasets    & Samples & Views & Clusters  \\ 
\hline
Prokaryotic \cite{brbic2016landscape}& 551     & 3     & 4         \\
Synthetic3d \cite{kumar2011co} & 600     & 3     & 3         \\
MNIST-USPS \cite{peng2019comic}  & 5000    & 2    & 10      \\
CCV  \cite{jiang2011consumer}       & 6773    & 3     & 20        \\
Hdigit \cite{Chem2022review}     & 10000   & 2     & 10        \\
Cifar10 \tablefootnote{http://www.cs.toronto.edu/kriz/cifar.html
}    & 50000   & 3     & 10        \\
Cifar100 \tablefootnote{http://www.cs.toronto.edu/kriz/cifar.html
}   & 50000   & 3     & 100       \\
YouTubeFace\tablefootnote{https://www.cs.tau.ac.il/ wolf/ytfaces/
}  & 101499  & 5     & 31        \\
Caltech-5V \cite{fei2004learning}  & 1400    & 5     & 7         \\
NGs \tablefootnote{https://lig-membres.imag.fr/grimal/data.html}  & 500  & 3     &5 \\
Cora \cite{wen2020generalized} & 2708   & 2     & 7\\
BDGP \cite{cai2012joint} & 2500   & 2     & 5\\ 
Fashion \cite{xiao2017novel} & 10000   & 3     & 10\\
\bottomrule
\end{tabular}
\label{tab:Datasets}
\end{table}
\begin{table*}[ht]
\renewcommand\arraystretch{1}
\centering
\caption{Clustering result comparison for different datasets.}
\resizebox{\textwidth}{!}{
\begin{tabular}{c|c|c|c|c|c|c|c|c|c|c|c|c} 
\hline
Datasets       & \multicolumn{3}{c|}{CCV}                           &\multicolumn{3}{c|}{ MNIST-USPS }          & \multicolumn{3}{c|}{Prokaryotic}                           & \multicolumn{3}{c}{Synthetic3d}                           \\ 
\hline
Metrics       & ACC           & NMI           & PUR           & ACC           & NMI           & PUR           & ACC           & NMI           & PUR           & ACC           & NMI           & PUR                          \\
\hline
PLCMF \cite{wang2021pseudo}& 0.2294 &0.1852 &0.2557          & 0.6228 &0.6594 &0.6670 & 0.4446 &0.0200 &0.5681                  & 0.8867 &0.6555 &0.8867                   \\
LMVSC \cite{kang2020large}& 0.2014 &0.1657 &0.2396          & 0.5626 &0.5039 &0.6060 & 0.5753 &0.1337 &0.6294 & 0.9567 &0.8307 &0.9567                   \\
SMVSC \cite{sun2021scalable}& 0.2182 &0.1684 &0.2439          &0.7542 &0.6883 &0.7542 & 0.5590 &0.1820 &0.5717                  & 0.9683 &0.8665 &0.9683                   \\
FastMICE \cite{huang2022fast} & 0.1997 &0.1518 &0.2341          & 0.9570 &0.9332 &0.9573 & 0.5629 &0.2685 &0.6500                  & 0.9613 &0.8490 &0.9613                   \\
DEMVC \cite{xu2021deep}    & 0.1942 &0.2113 &0.2169          &  0.8858 &0.9100 &0.8880 & 0.5245 &0.3079 &0.6969          & 0.8100 &0.6136 &0.8100                       \\
CONAN \cite{ke2021conan}    & 0.1422 &0.1016 &0.1674          & 0.5722 &0.5708 &0.6178 & 0.4809 &0.1589 &0.5045                  & 0.9650 &0.8540 &0.9650                     \\
SiMVC \cite{trosten2021reconsidering}    & 0.1513 &0.1252 &0.2161    & 0.9810 &0.9620 &0.9810  & 0.5009 &0.1945 &0.6098                  & 0.9366 &0.7747 &0.9366                   \\
CoMVC \cite{trosten2021reconsidering}   & 0.2962 &0.2865 &0.2976  & 0.9870 &0.9760 &0.9890 & 0.4138 &0.1883 &0.6697                  & 0.9530 &0.8184 &0.9520                       \\
MFLVC \cite{xu2022multi}    & 0.3123 &0.3162 &0.3391             & 0.9954 &0.9869 &0.9898    & 0.4301 &0.2216 &0.5989                  & 0.9650 &0.8537 &0.9650                   \\
Ours    & \textbf{0.3543} & \textbf{0.3292} & \textbf{0.3812} & \textbf{0.9956} &\textbf{0.9871} &\textbf{0.9956} & \textbf{0.6225} &\textbf{0.3778} &\textbf{0.7314}        & \textbf{0.9700} & \textbf{0.8713} & \textbf{0.9700}  \\
\hline
\end{tabular}}
\label{tab:Clustering performance1}
\end{table*}
\begin{table*}[ht]
\renewcommand\arraystretch{1}
\centering
\caption{Clustering result comparison for different datasets.}
\resizebox{\textwidth}{!}{
\begin{tabular}{c|c|c|c|c|c|c|c|c|c|c|c|c} 
\hline
Datasets       & \multicolumn{3}{c|}{Hdigit}                           &\multicolumn{3}{c|}{ YouTubeFace }          & \multicolumn{3}{c|}{Cifar10}                           & \multicolumn{3}{c}{Cifar100 }                           \\
\hline                        
Metrics       & ACC           & NMI           & PUR           & ACC           & NMI           & PUR           & ACC           & NMI           & PUR           & ACC           & NMI           & PUR                          \\
\hline
PLCMF \cite{wang2021pseudo}   & 0.9047 &0.7965 &0.9047                          & 0.1473 &0.1237 &0.2875   & 0.8144 &0.8265 &0.8497   & 0.8260 &0.9593 &0.8698          \\
LMVSC \cite{kang2020large}    & 0.9709 &0.9293 &0.9709                  & 0.1479 &0.1327 &0.2816   & 0.9896 &0.9721 &0.9896   & 0.8482 &0.9583 &0.9582          \\
SMVSC \cite{sun2021scalable}    & 0.8634 &0.7683 &0.8634                          & 0.2587 &0.2292 &0.3321   & 0.9899 &0.9730 &0.9899   & 0.7429 &0.9091 &0.7529          \\
FastMICE \cite{huang2022fast} &0.9332 &0.9258 &0.9417                          & 0.1825 &0.1633 &0.3028   & 0.9694 &0.9622 &0.9704   & 0.8257 &0.9464 &0.8298          \\
DEMVC \cite{xu2021deep}    & 0.3738 &0.3255 &0.4816                          & 0.2487 &0.0932 &0.2662   & 0.4354 &0.3664 &0.4498   & 0.5048 &0.8343 &0.5177          \\
CONAN \cite{ke2021conan}    & 0.9562 &0.9193 &0.9562                          & 0.1179 &0.1178 &0.1499  & 0.9255 &0.8641 &0.9255   & 0.6711 &0.9441 &0.9983  \\
SiMVC \cite{trosten2021reconsidering}    & 0.7854 &0.6705 &0.7854                          & 0.0765 &0.0481 &0.2662 & 0.8359 &0.7324 &0.8359 & 0.5795 &0.9225 &0.5869        \\
CoMVC \cite{trosten2021reconsidering}    & 0.9032 &0.8713 &0.9032                          & 0.1010 &0.0851 &0.2674   & 0.9275 &0.8925 &0.9275   & 0.6569 &0.9345 &0.6570          \\
MFLVC \cite{xu2022multi}    & 0.9442 &0.8750 &0.9440                             & 
0.2770 &0.2952 &0.3297   & 0.9918 &0.9774 &0.9918   & 0.8268 &0.9560 &0.8268          \\
Ours    & \textbf{0.9744} &\textbf{0.9305} &\textbf{0.9744} & \textbf{0.3262} &\textbf{0.3289} &\textbf{0.4007}   &\textbf{0.9923} &\textbf{0.9781} &\textbf{0.9923}   & \textbf{0.9597} &\textbf{0.9935} &\textbf{0.9605}          \\
\hline
\end{tabular}}
\label{tab:Clustering performance2}
\end{table*}
\begin{table*}[ht]
\renewcommand\arraystretch{1}
\centering
\caption{Comparisons with deep clustering methods on Caltech dataset with increased views. “XV” denotes the number of views.}   
\resizebox{\textwidth}{!}{
\begin{tabular}{c|c|c|c|c|c|c|c|c|c|c|c|c} 
\hline
Datasets       & \multicolumn{3}{c|}{Caltech-2V }                           &\multicolumn{3}{c|}{  Caltech-3V  }          & \multicolumn{3}{c|}{Caltech-4V}                           & \multicolumn{3}{c}{Caltech-5V}                           \\ 
\hline
Metrics       & ACC           & NMI           & PUR           & ACC           & NMI           & PUR           & ACC           & NMI           & PUR           & ACC           & NMI           & PUR                          \\
\hline
DEMVC \cite{xu2021deep}    & 0.4986 &0.3845 &0.5207                   & 0.5336 &0.4136 &0.5336                 & 0.4929 &0.4504 &0.5186 & 0.4600 &0.3666 &0.4950  \\
CONAN \cite{ke2021conan}    & 0.5750 &0.4516 &0.5757                    & 0.5914 &0.4981 &0.5914                 & 0.5571 &0.5061 &0.5735 & 0.7207 &0.6418 &0.7221  \\
SiMVC \cite{trosten2021reconsidering}    & 0.5083 &0.4715 &0.5573                      & 0.5692 &0.4953 &0.5912                    & 0.6193 &0.5362 &0.6303    & 0.7193 &0.6771 &0.7292     \\
CoMVC \cite{trosten2021reconsidering}    & 0.4663 &0.4262 &0.5272                      & 0.5413 &0.5043 &0.5842                    & 0.5683 &0.5692 &0.6463    & 0.7003 &0.6871 &0.7462     \\
MFLVC \cite{xu2022multi}    & 0.6060 &\textbf{0.5280} &0.6160     & 0.6312 &\textbf{0.5663} &0.6392                    & 0.7332 &0.6523 &0.7342    & 0.8042 &0.7032 &0.8043     \\
Ours  &  \textbf{0.6643} &0.5008 &\textbf{0.6643}    &  \textbf{0.6400} &0.5345 &\textbf{0.6529} & \textbf{0.7343} &\textbf{0.6610} &\textbf{0.7343}  &\textbf{0.8336} &\textbf{0.7331} &\textbf{0.8336}  \\
\hline
\end{tabular}}
\label{tab:Clustering performance3}
\end{table*}

\textbf{Compared methods:} 
To evaluate the effectiveness of the proposed method, we compare the proposed method with 13 SOTA clustering methods, including 9 methods for complete multi-view datasets and 4 methods for incomplete multi-view datasets. 
The former includes 4 traditional methods (PLCMF \cite{wang2021pseudo}, LMVSC \cite{kang2020large}, SMVSC \cite{sun2021scalable}, and
FastMICE \cite{huang2022fast}) and 5 deep methods (DEMVC \cite{xu2021deep},
CONAN \cite{ke2021conan}, SiMVC \cite{trosten2021reconsidering}, CoMVC \cite{trosten2021reconsidering}, MFLVC \cite{xu2022multi}).
The latter includes CDIMC  \cite{ijcai2020p447}, 
 COMPLETER \cite{lin2021completer}, DIMVC\cite{xu2022deep}, DSIMVC \cite{tang2022deep}.


\begin{table*}
\centering
\caption{Clustering results on incomplete datasets.  "-" denotes unknown results as COMPLETER mainly focuses on two-view clustering.}
\resizebox{\textwidth}{!}{
\begin{tabular}{c|c|c|c|c|c|c|c|c|c|c|c|c|c} 
\hline
\multirow{2}{*}{Dataset}     & Missing rates      & \multicolumn{3}{c|}{0.1}                      & \multicolumn{3}{c|}{0.3}                      & \multicolumn{3}{c|}{0.5}                      & \multicolumn{3}{c}{0.7}                        \\ 
\cline{2-14}
                             & Evaluation metrics & ACC           & NMI           & PUR           & ACC           & NMI           & PUR           & ACC           & NMI           & PUR           & ACC           & NMI           & PUR            \\ 
\hline
\multirow{5}{*}{\rotatebox{90}{BDGP}}        & CDIMC\cite{ijcai2020p447}          & 0.8047         & 0.7008         & 0.8037         & 0.7467         & 0.6764         & 0.7527         & 0.6771         & 0.5451         & 0.6771         & 0.5611         & 0.3970          & 0.5776          \\
                             & COMPLETER\cite{lin2021completer}          & 0.4091         & 0.4180          & 0.4154         & 0.3963         & 0.3319         & 0.3115         & 0.3262         & 0.2747         & 0.4390          & 0.4359         & 0.4510          & 0.4090           \\
                             & DIMVC\cite{xu2022deep}              & 0.9640         & 0.8920         & 0.9120         & 0.9540         & 0.8660         & 0.8890         & 0.9470         & 0.8450         & 0.8730         & 0.9290         & 0.8020         & 0.8310          \\
                             & DSIMVC\cite{tang2022deep}             & 0.9827         & 0.9443         & 0.9827         & 0.9693         & 0.9034         & 0.9693         & 0.9529         & 0.8611         & 0.9529         & 0.9214         & 0.7937         & 0.9214          \\
                             & DSIMVC++    (Our)       & \textbf{0.9836}        & \textbf{0.9455}        & \textbf{0.9836}        & \textbf{0.9698}        & \textbf{0.9050}         & \textbf{0.9698}        & \textbf{0.9557}        & \textbf{0.8685}        & \textbf{0.9557}        & \textbf{0.9332}        & \textbf{0.8142}        & \textbf{0.9332}         \\ 
\hline
\multirow{5}{*}{\rotatebox{90}{Synthetic3d}} & CDIMC\cite{ijcai2020p447}          & 0.5965        & 0.3564        & 0.6000        & 0.5124        & 0.2239        & 0.5340        & 0.5330        & 0.2373        & 0.5663        & 0.4136        & 0.1387        & 0.4394         \\
                             & COMPLETER\cite{lin2021completer}          & -           &        -       &       -        &     -          &               &       -        &      -         &         -      &          -     &         -      &               &     -           \\
                             & DIMVC\cite{xu2022deep}               & \textbf{0.8183}   & 0.6701   & 0.8380   & \textbf{0.8233}   & 0.5860  & 0.8241   & \textbf{0.7968}  & 0.5355   & 0.7971   & 0.6689  & 0.3974  & 0.6774  \\
                             & DSIMVC \cite{tang2022deep}            & 0.7613& 0.6744 & 0.8943 & 0.7378 & 0.6365  & 0.8773  & 0.7247  & 0.6090 & 0.8643  & 0.7043 & 0.5499 & 0.8242  \\
                             & DSIMVC++  (Our)         & 0.7785 & \textbf{0.6933} & \textbf{0.9005} & 0.7530  & \textbf{0.6693} & \textbf{0.9042} & 0.7612 & \textbf{0.6463} & \textbf{0.8952}  & \textbf{ 0.7197} & \textbf{ 0.5900}  & \textbf{ 0.8638}  \\ 
\hline

\multirow{5}{*}{\rotatebox{90}{Cora}}        & CDIMC\cite{ijcai2020p447}          & 0.2460 & 0.0111 & 0.3066 & 0.2222 & 0.0066 & 0.3024 & 0.2400 & 0.0052& 0.3022 & 0.2518 & 0.0054 & 0.3025  \\
                             & COMPLETER \cite{lin2021completer}          & 0.2441    & \textbf{0.4300}     & 0.3172   & 0.2542  & \textbf{0.4130}    & 0.3242   & 0.2464  & \textbf{0.4070}   & 0.3199   & 0.2540   & 0.1850    & 0.3055    \\
                             & DIMVC     \cite{xu2022deep}          & 0.4384        & 0.2231        & 0.5079        & 0.3704        & 0.1470         & 0.4082        & 0.3561        & 0.1432        & 0.4275        & 0.2789        & 0.0718        & 0.3397         \\
                             & DSIMVC   \cite{tang2022deep}          & 0.4402   & 0.3316  & 0.5445   & 0.4106   & 0.2924   & 0.5099  & 0.3764  & 0.2360  & 0.4742   & 0.3243  & 0.1628   & 0.4228 \\
                             & DSIMVC++ (Our)          & \textbf{0.4699}   & 0.3271  & \textbf{0.5588 } & \textbf{0.4484}  & 0.3035  & \textbf{0.5544}  & \textbf{0.4338} & 0.2720  & \textbf{0.5290}& \textbf{0.3554 } & \textbf{0.1935}   & \textbf{0.4620}  \\ 
\hline
\multirow{5}{*}{\rotatebox{90}{NGs}}        & CDIMC \cite{ijcai2020p447}         & 0.3072 & 0.0794& 0.3216& 0.2736 & 0.0478& 0.2832& 0.2532 & 0.0346 & 0.2620& 0.2464 & 0.0270 & 0.2504 \\
                             & COMPLETER  \cite{lin2021completer}        & -             &  -             &   -            &      -         &   -            &    -           &     -          &        -       &       -        &     -          &   -            &    -            \\
                             & DIMVC \cite{xu2022deep}             & 0.3543        & 0.1493        & 0.3562        & 0.2120         & 0.0363        & 0.2138        & 0.2213        & 0.0588        & 0.2280         & 0.2598        & 0.0546        & 0.2645         \\
                             & DSIMVC  \cite{tang2022deep}           & 0.5564 & 0.4599& 0.6230 & 0.5178  & 0.3864 & 0.5854& \textbf{0.4672}& \textbf{0.2980} & 0.5244 & 0.4090& 0.2095& 0.4746\\
                             & DSIMVC++ (Our)          & \textbf{0.6358} &\textbf{ 0.5186} & \textbf{0.7090}& \textbf{0.6136}& \textbf{0.4428} & \textbf{0.6734} & 0.4598 &0.2801 &\textbf{ 0.5310} &\textbf{ 0.4410 }& \textbf{0.2298}&\textbf{ 0.5054} \\ 
\hline
\multirow{5}{*}{\rotatebox{90}{Fashion}}     & CDIMC\cite{ijcai2020p447}          & 0.6500 & 0.6642 & 0.6696 & 0.5064& 0.5121 & 0.5241& 0.4484& 0.4483& 0.4553& 0.3693 & 0.3668 & 0.3818 \\
                             & COMPLETER\cite{lin2021completer}          &  -           &    -           &       -        &        -       &         -      &       -        &           -    &           -    &      -         &       -        &     -          &       -         \\
                             & DIMVC \cite{xu2022deep}             & 0.7811        & 0.8578        & 0.8286        & 0.7132        & 0.7676        & 0.7614        & 0.7044        & 0.7447        & 0.7508        & 0.6128        & 0.6806        & 0.6693         \\
                             & DSIMVC \cite{tang2022deep}            & 0.8798  & 0.8623  & 0.8800  & 0.8680  & 0.8379  & 0.8687  & 0.8333 & 0.8025  & 0.8337  & 0.7825 & 0.7626 & 0.7825   \\
                             & DSIMVC++ (Our)          & \textbf{0.9360}  & \textbf{0.8953}  & \textbf{0.9360} & \textbf{0.9160} & \textbf{0.8657} & \textbf{0.9160}   & \textbf{0.8969 } &\textbf{ 0.8366}  & \textbf{0.8969} &\textbf{ 0.8637 } & \textbf{0.8015}  &\textbf{ 0.8644}  \\ 
\hline
\end{tabular}}
\label{tab:incomplete performance}
\end{table*}

\textbf{Implementation Details:} 
The experiments were conducted on Linux with i9-10900K CPU, 62.5GB RAM and 3090Ti GPU. We pre-train the models for 200 epochs for the reconstruction loss and then fine-tune the model for 100 epochs on mini-batches of size 256 using the Adam optimizer \cite{kingma2014adam} in the PyTorch \cite{paszke2019pytorch} framework. 
In the proposed method, we reshape all datasets into vectors and implement autoencoders using a fully connected network. 

For the comparison with these methods on incomplete multi-view datasets, the incomplete samples is ready according to the method \cite{tang2022deep} on Synthetic3D, NGs, Cora, BDGP, and Fashion by randomly removing views under the condition that at least one view remained in the sample. The ratio of incomplete sample sizes to overall sample sizes is set from 0.1 to 0.7 with 0.2 as the interval. 
Please refer to the supplementary material for the experiment details for the proposed method in incomplete datasets. 
\subsection{Experimental comparative results}
The comparative results with 9 methods by three evaluation metrics (ACC, NMI, PUR) on 8 benchmark datasets with different scales are presented as Table~\ref{tab:Clustering performance1} and Table~\ref{tab:Clustering performance2}. 
From the tables, we can observe that the proposed GCFAggMVC obtains better results than those of other methods. Specifically, we obtain the following observations:

(1)  Our method obtains better results than four traditional multiview clustering methods (PLCMF, LMVSC, SMVSC, and FastMICE). These methods learn the data self-representation or graph structure relationship on the raw data features. These raw data features contain noises and redundancy information, which is harmful to establish the essential graph structure.  From this table, it can be seen that our method achieves better performance.

(2) We compare five deep multiview clustering methods  (DEMVC, CONAN, SiMVC, CoMVC,  and MFLVC) with the proposed method. In DEMVC,  they consider to align  label distributions of different views to a target distribution, which has a negative impact since a given cluster distribution from one view might be aligned with a different cluster distribution from another view~\cite{trosten2021reconsidering}.  In CONAN, SiMVC, and CoMVC, they obtain a consensus feature representation by view-wise fusion method. However, the meaningless view-private information might be dominant in the feature fusion, and thus harmful to the clustering performance. Compared with these methods, in the proposed method, the consensus representation of each sample from different views can be enhanced by those samples with high structure relationship, moreover, maximizes the consistency of the view-specific representations and consensus data representation to obtain efficient clustering performance.  From this table, it can be seen that GCFAggMVC obtains better results than those of other methods.

To further verify the effectiveness of the proposed method, we test the performance of different numbers of views on the Caltech dataset. Table \ref{tab:Clustering performance3} shows the comparative results with different deep clustering methods. From this table, we can see that the results of the proposed GCFAggMVC outperform those of the other methods. 

Since our consensus representation is enhanced by those samples with high structure relationships, the proposed GCFAgg module and SgCL loss canbe plugged into incomplete multi-view clustering task. To verify the effectiveness, we integrate the proposed GCFAgg module and SgCL loss to the state-of-the-art DSIMVC (which is an incomplete multi-view clustering method) framework (named DSIMVC++, please refer
to the appendix for the details) and compare it with the other 4 methods in Table \ref{tab:incomplete performance}. In the DSIMVC++ method, the consensus feature presentation for clustering can be enhanced by those samples with high structure relationship,  hence, the propose method can better cope with incomplete multiview clustering tasks.  From the table, it can be observed that our GCFAggMVC  achieves better performance than other methods. Especially, when the missing view rate is 0.7, it can be observed that the proposed GCFAggMVC  can obtain the best results than other all comparison methods.

\subsection{Model analysis}
\begin{figure*}[ht]
    \centering
     \subfloat[]{
       \includegraphics[width=0.255\linewidth]{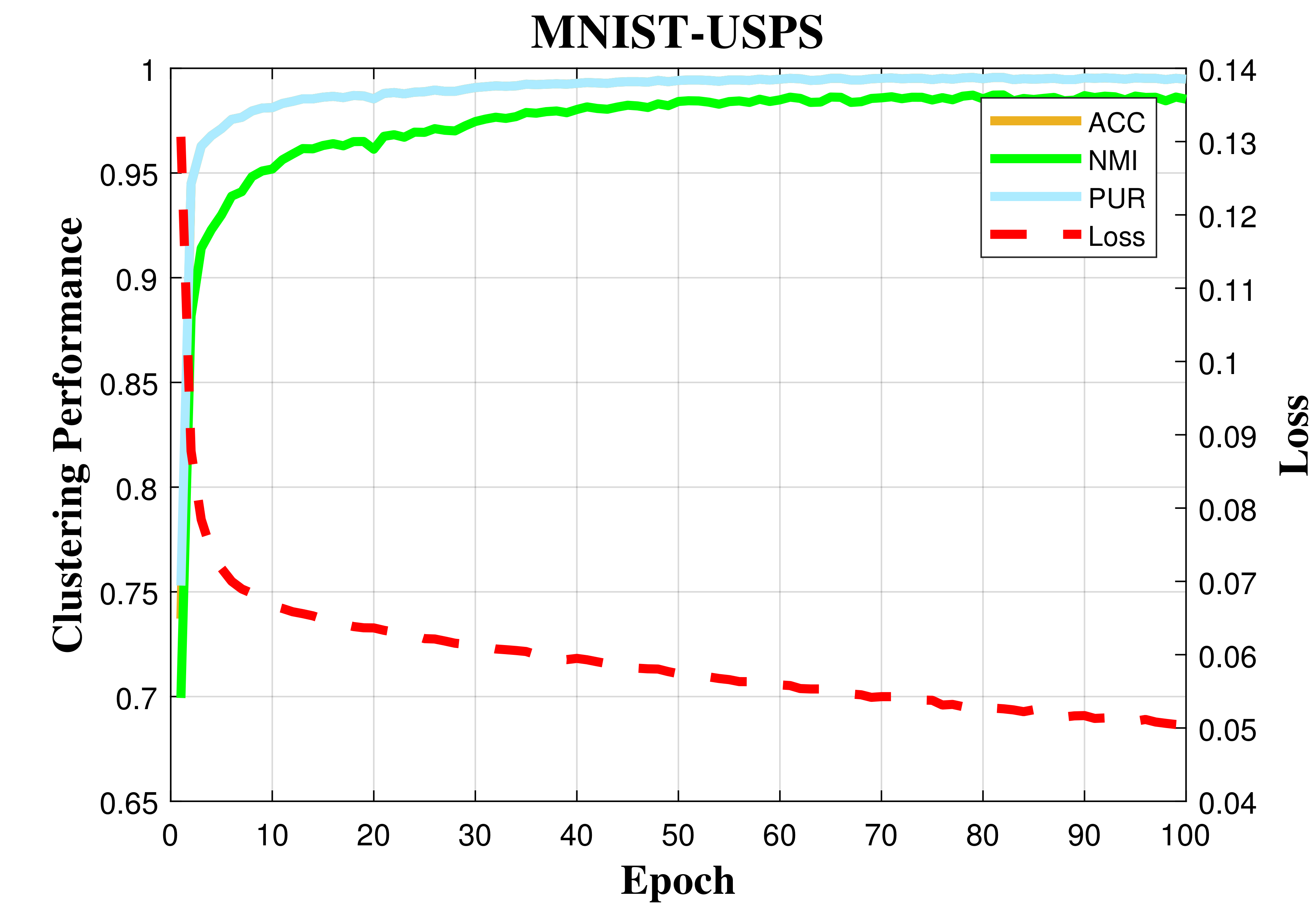}}
	  \subfloat[]{
        \includegraphics[width=0.25\linewidth]{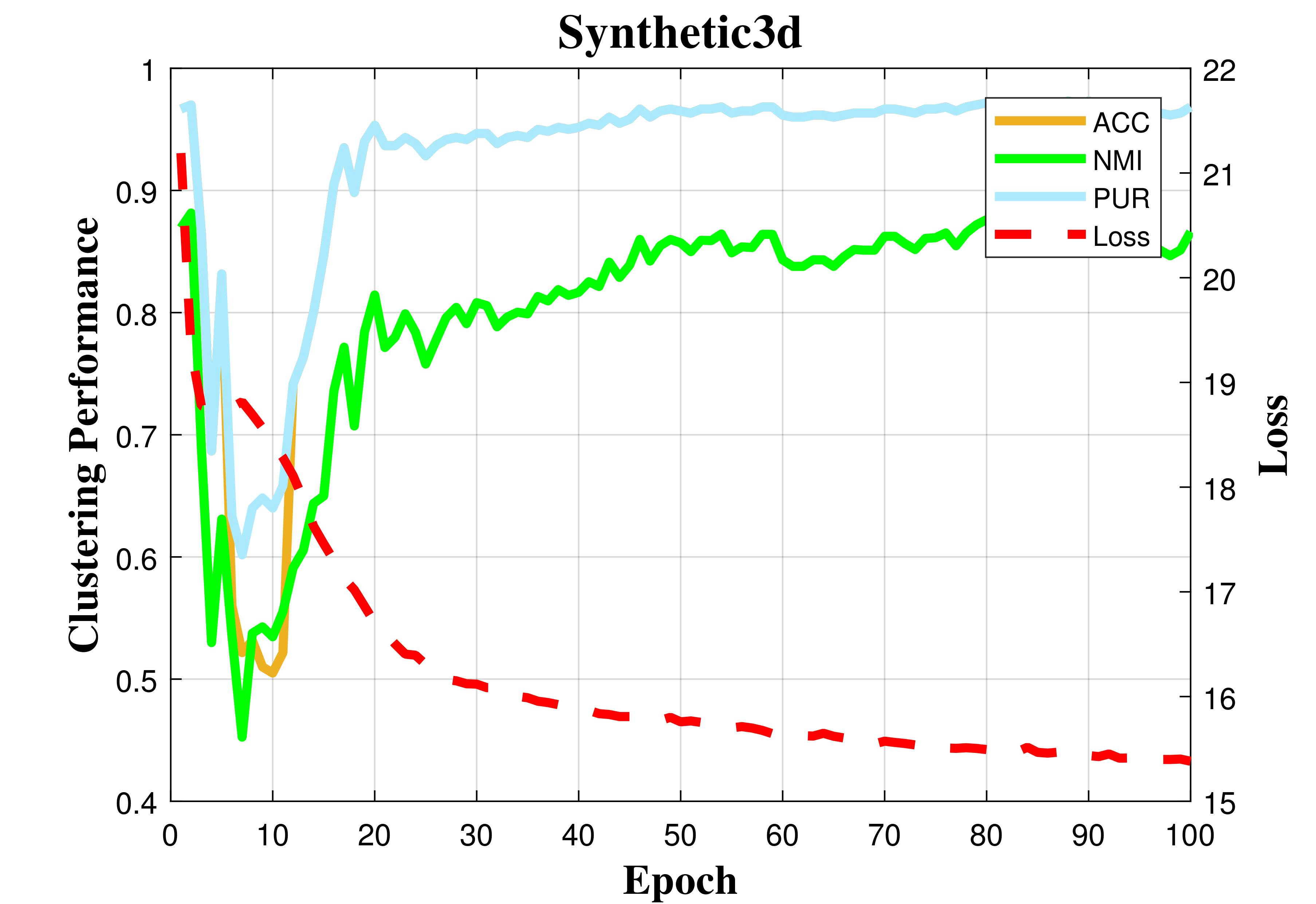}}
         \subfloat[]{
        \includegraphics[width=0.25\linewidth]{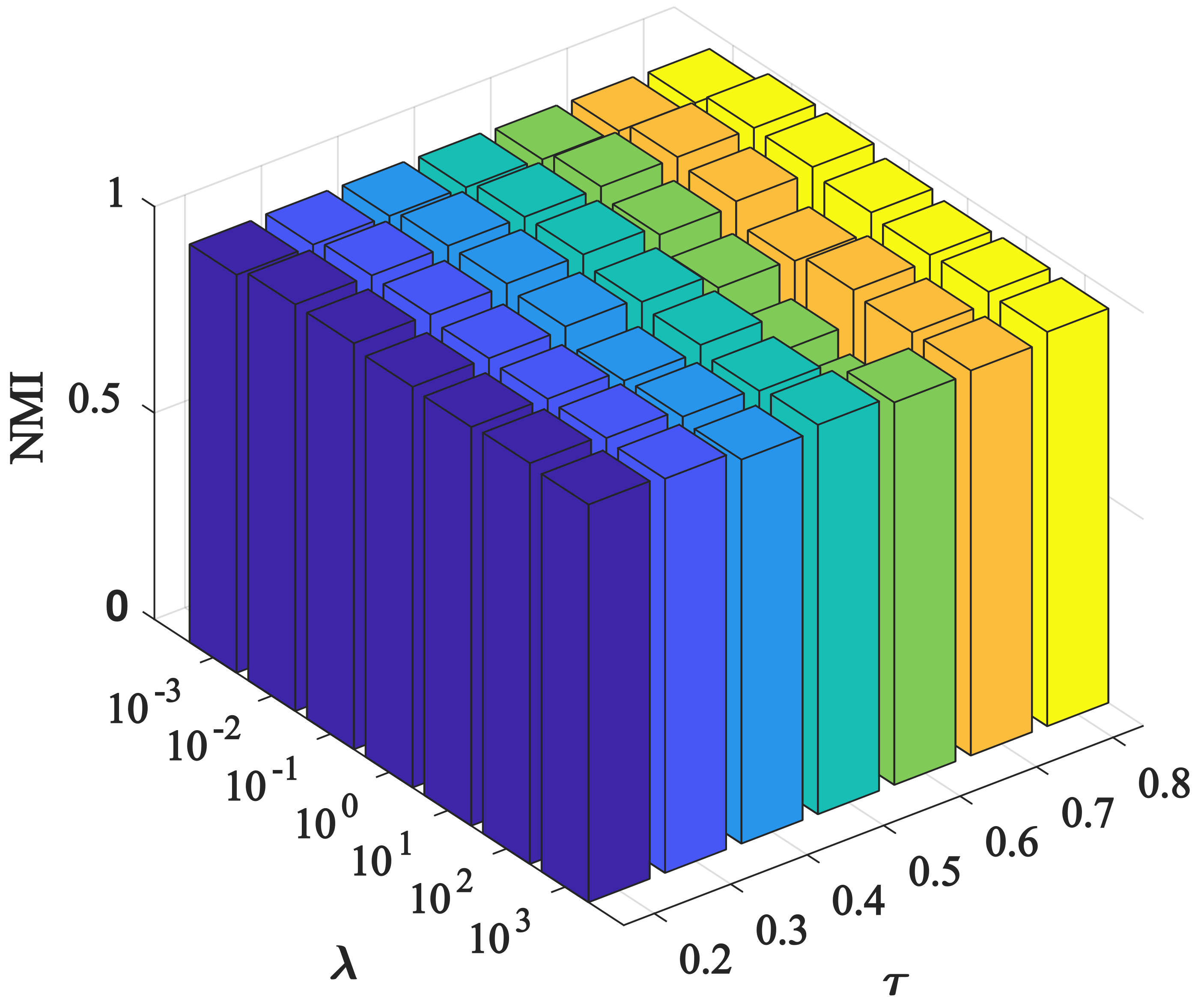}}
                 \subfloat[]{
        \includegraphics[width=0.25\linewidth]{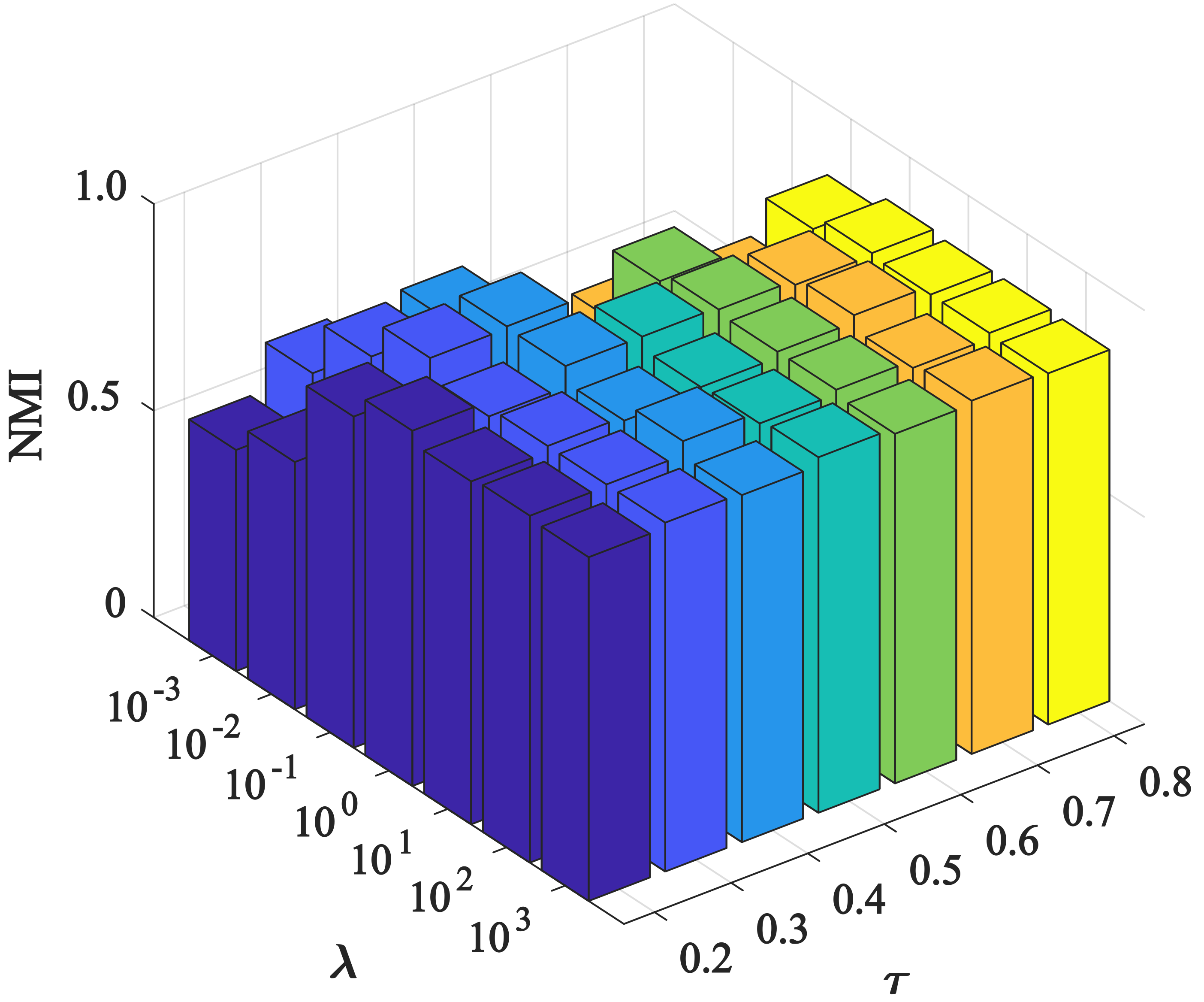}}
	  \caption{The  convergence analysis and parameter analysis on  MNIST-USPS, Synthetic3d, respectively.}
	  \label{fig:Visualization and convergence} 
\end{figure*}

\textbf{Convergence analysis.} 
To verify the convergence, we plot the objective values and evaluation metric values through iterations in Figure \ref{fig:Visualization and convergence}. It can be observed that the objective value monotonically decreases until convergence. The value of ACC, NMI, and PUR first increases gradually with iteration and then fluctuate in a narrow range. These results all confirm the convergence of GCFAggMVC.


\textbf{Parameter sensitivity analysis.} We experimentally evaluate the effect of hyperparameters on the clustering performance of GCFAggMVC, which includes the trade-off coefficient $\lambda$ (i.e., $\mathcal{L}_{\mathrm{r}}+\lambda \mathcal{L}_{\mathrm{c}}$) and the temperature parameter $\tau$.  Figure \ref{fig:Visualization and convergence} shows the NMI of GCFAggMVC when $\lambda$ is varied from $10^{-3}$ to $10^{3}$ and $\tau$ from 0.2 to 0.8. From this figure, the clustering results of the proposed GCFAggMVC  are insensitive to both $\lambda$
and $\tau$ in the range 0.1 to 1, and the range 0.3 to 0.5, respectively. 
Empirically, we set $\lambda$ and $\tau$ to 1.0 and 0.5.

\begin{table}[tb]
\centering
\caption{Ablation study.}
\resizebox{\textwidth/2}{!}{
\begin{tabular}{ccccc} 
\hline
Datasets                     & Method      & ACC & NMI  & PUR \\ 
\hline
\multirow{3}{*}{Prokaryotic} & No-GCFAgg & 0.4403 & 0.1906  & 0.5740 \\
                             & No-SgCL & 0.4804 & 0.2226 & 0.6534 \\
                             & GCFAggMVC       & \textbf{0.6225} & \textbf{0.3778} & \textbf{0.7314}  \\
\hline
\multirow{3}{*}{CCV}         & No-GCFAgg & 0.2850  & 0.2740 & 0.3150 \\
                             & No-SgCL & 0.2020 & 0.1900 & 0.2560            \\
                             & GCFAggMVC & \textbf{0.3543}  & \textbf{0.3292}& \textbf{0.3812}   \\ 
\hline
\multirow{3}{*}{MNIST-USPS} & No-GCFAgg & 0.9753 & 0.9500  & 0.9753 \\
                             &  No-SgCL & 0.6949 & 0.6656 & 0.7410 \\
                             & GCFAggMVC        & \textbf{0.9956} & \textbf{0.9871} & \textbf{0.9956}  \\
\hline
\end{tabular}}
\label{tab:Ablation}
\end{table}
\begin{table}[h]
\centering
\caption{The ablation study for the SgCL.}
\begin{tabular}{ccccc} 
\hline
Datasets                     & Method      & ACC & NMI  & PUR \\ 
\hline
\multirow{3}{*}{CCV} & Standard CL & 0.2711 & 0.2669  & 0.3046 \\
                             & Standard CL with $\mathbf{S}$ & 0.3046 & 0.3017 & 0.3363 \\
                             & SgCL without $\mathbf{S}$ & 0.2858 & 0.2833 & 0.3260 \\
                             & SgCL      & \textbf{0.3543} & \textbf{0.3292} & \textbf{0.3812}  \\
\hline
\multirow{3}{*}{MNIST-}        & Standard CL & 0.9562 & 0.9386  & 0.9562 \\
\multirow{3}{*}{USPS}                           & Standard CL with $\mathbf{S}$ & 0.9768 & 0.9527& 0.9768 \\
                             & SgCL without $\mathbf{S}$ & 0.9698 & 0.9327 & 0.9698 \\
                             & SgCL      & \textbf{0.9956} & \textbf{0.9871} & \textbf{0.9956}  \\
\hline
\end{tabular}
\label{tab:Ablation1}
\end{table}

\begin{figure}[!ht] 
    \centering
	  \subfloat[ $\mathbf{Z}$ features]{
       \includegraphics[width=0.33\linewidth]{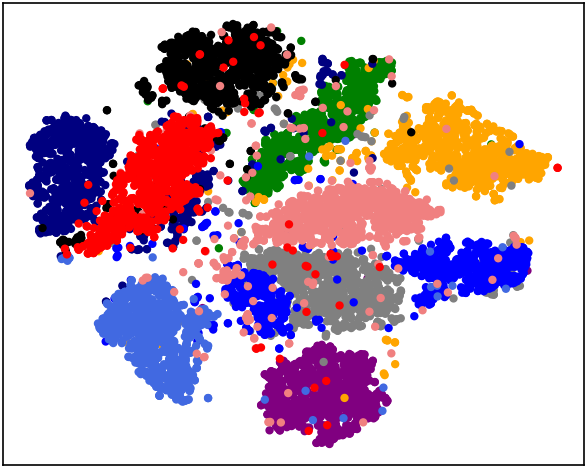}}
	  \subfloat[ $\mathbf{H}$ features]{
        \includegraphics[width=0.33\linewidth]{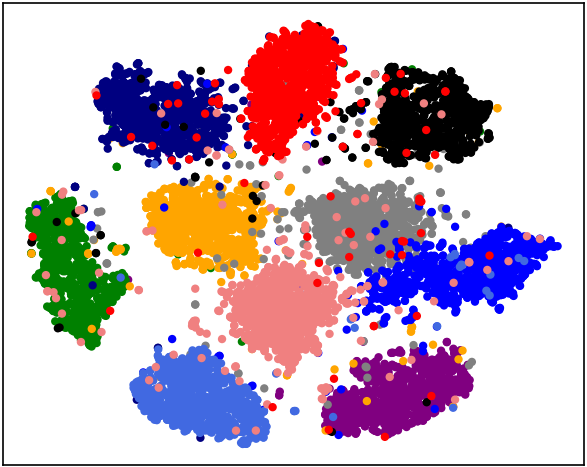}}
	  \subfloat[ $\hat{\mathbf{H}}$ features]{
        \includegraphics[width=0.33\linewidth]{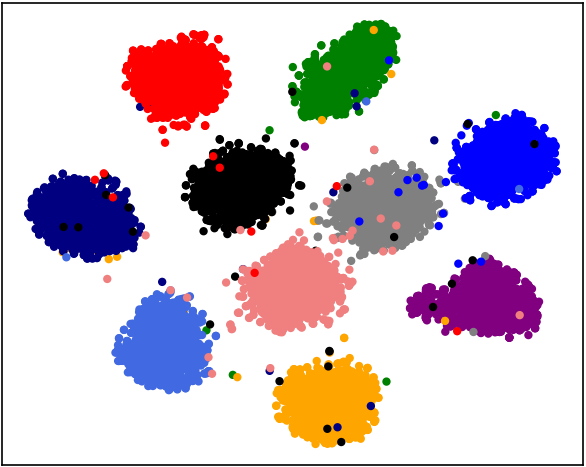}}
	  \caption{The visualization results of different  feature representations on different layers after convergence. Note that, $\mathbf{H}$ feature denotes the concatenation of all learnt $\mathbf{H}^v$. }
	  \label{fig:t-SNE} 
\end{figure}



\par
\textbf{Ablation Study.} We conducted an ablation study to evaluate
each component of the proposed model. We perform two models and compare them with the proposed method.

\par \textit{Validity of the proposed GCFAgg module}: we set the   $\mathbf{Z}$  (concatenation of all view-specific representations)  as  the consensus representation. This method is denoted as "No-GCFAgg".  As
shown in Table \ref{tab:Ablation}, it can be observed that the results of No-GCFAgg are lower than the results of the proposed method by 18.22, 6.93, and 2.03 percent  in ACC term.  The concatenated representation   $\mathbf{Z}$  includes much view-private information, which is not conductive to clustering. The GCFAgg fully explores the complementary of similar samples, thereby reducing the impact of noise and redundancy among all views.  The results show that our GCFAgg module can 
enhance the consistency of data representation from the same cluster.
\par \textit{Effectiveness of the proposed SgCL loss}: 
As shown in Table \ref{tab:Ablation}, the results of No-SgCL  are lower than those of the proposed method by 14.21, 15.23, and 30.07 percent  in ACC term. 
 Since our consensus representation of multiple views is enhanced by global structure relationship of all samples, and moreover the contrastive learning maximizes the similarity of the consensus representation and view-specific representation from the same sample,  minimize the similarity of the representations from different samples with low structure relationship, which improves the clustering performance. More analysis results for SgCL are shown in the supplementary material.
 
 In the experiment, we set the contrastive learning with the sample-level loss as Standard CL. That is, these inter-view presentations from the same sample are set as positive pairs, and  view representations from different samples are set as negative pairs (such as the contrastive learning in \cite{tang2022deep,xu2022multi} ). The ablation study for the SgCL is show in Table \ref{tab:Ablation1}. The experiment shows the effectiveness of our SgCL compared with the standard CL.
 
In addition, to further verify the effectiveness of the proposed GCFAggMVC,  we visualize different feature representations on different layers after convergence by the t- SNE method \cite{van2008visualizing} in Figure \ref{fig:t-SNE}. It can be clearly observed that the clustering structure of our learnt consensus features  $\widehat {\bf{H}}$ is clearer than that of the concatenation of view-specific representation $\bf{Z}$  and view-specific representation $\bf{H}$, and  yields well-separated clusters. 
\vspace{-2mm}
\section{Conclusion}
In this paper, we propose a novel multi-view representation learning network for clustering. We first learn the view-specific features to reconstruct the original data by leveraging the autoencoder model. And then, we design a GCFAgg module, which can learn a global similarity relationship among samples, and moreover enhance a consensus representation by all samples’ representation, which makes the representations of these samples with high structure relationship be more similar. Furthermore, we design the SgCL module, which addresses the problem that the representations from different samples in the same cluster is inconsistent in previous contrastive learning methods. Extensive experimental results show that our approach has SOTA performance in both complete and incomplete MVC tasks.
\vspace{-4mm}
\section*{Acknowledgments}
This work was partially supported by the National Natural Science Foundation of China (Grants No. 61801414, No. 62076228, and No. 62001302).   \\
{\small
\bibliographystyle{ieee_fullname}
\bibliography{egbib}
}

\end{document}